\pdfoutput=1

\documentclass[11pt]{article}

\usepackage[final]{acl}
\usepackage{times}
\usepackage{latexsym}
\usepackage[T1]{fontenc}
\usepackage[utf8]{inputenc}
\usepackage{microtype}
\usepackage{inconsolata}
\usepackage{graphicx}
\usepackage{tabularx}
\usepackage{subfigure}
\usepackage{subcaption}
\usepackage{amsmath}
\usepackage{multirow}
\usepackage{float}
\usepackage{xcolor}
\usepackage{booktabs}
\usepackage{makecell}
\usepackage{listings}

\usepackage[colorinlistoftodos]{todonotes}

\title{Society of Medical Simplifiers}

\author{Chen Lyu \\
  University of Warwick \\
  \texttt{chen.lyu@warwick.ac.uk} \\\And
  Gabriele Pergola \\
  University of Warwick \\
  \texttt{gabriele.pergola.1@warwick.ac.uk} \\}

\begin{document}
\maketitle
\begin{abstract}
Medical text simplification is crucial for making complex biomedical literature more comprehensible to non-experts. Traditional methods struggle with the specialized terms and jargon of medical texts, lacking the flexibility to adapt the simplification process dynamically.  In contrast, recent advancements in large language models (LLMs) present unique opportunities by offering enhanced control over text simplification through iterative refinement and collaboration between specialized agents. In this work, we introduce the \textit{Society of Medical Simplifiers}, a novel LLM-based framework inspired by the "Society of Mind" (SOM) philosophy. Our approach leverages the strengths of LLMs by assigning five distinct roles, i.e., Layperson, Simplifier, Medical Expert, Language Clarifier, and Redundancy Checker—organized into interaction loops. This structure allows the agents to progressively improve text simplification while maintaining the complexity and accuracy of the original content. Evaluations on the Cochrane text simplification dataset demonstrate that our framework is on par with or outperforms state-of-the-art methods, achieving superior readability and content preservation through controlled simplification processes.
\end{abstract}

\section{Introduction}
Medical text simplification is critical for improving public understanding of biomedical literature, which is often written in highly specialized language laden with domain-specific terminology that can be difficult for non-experts to understand \cite{WHO, devaraj2021paragraph, tsar-2022-text, lu2023napss}. Automatic text simplification (ATS) offers a potential solution by transforming complex biomedical language into simpler, more comprehensible text while preserving essential details. However, traditional approaches to medical text simplification often struggle to effectively handle the specialized terminology and syntactic complexity in medical literature, and lack effective mechanisms to granularly control the simplification process \cite{lu-etal-2023-napss, sun-etal-2022-phee}.

The recent advancements in large language models (LLMs) have revolutionized the ability to tackle complex tasks \cite{pergola-etal-2021-boosting, sun-etal-2022-phee, sun-etal-2024-leveraging}, opening new possibilities for refining ATS. What sets LLMs apart is their flexibility via prompt-engineering, which allows for fine-tuned control over the simplification process—an unprecedented capability in ATS. For the first time, LLMs can be specialized into distinct agents that collaborate and interact with each other in an iterative manner, dynamically refining the text simplification. This flexibility enables the simplification process to become more adaptive and progressive through multiple iterations, ensuring that both clarity and accuracy are maintained.

\begin{figure*}[ht]
\centering
\includegraphics[width=0.95\textwidth]{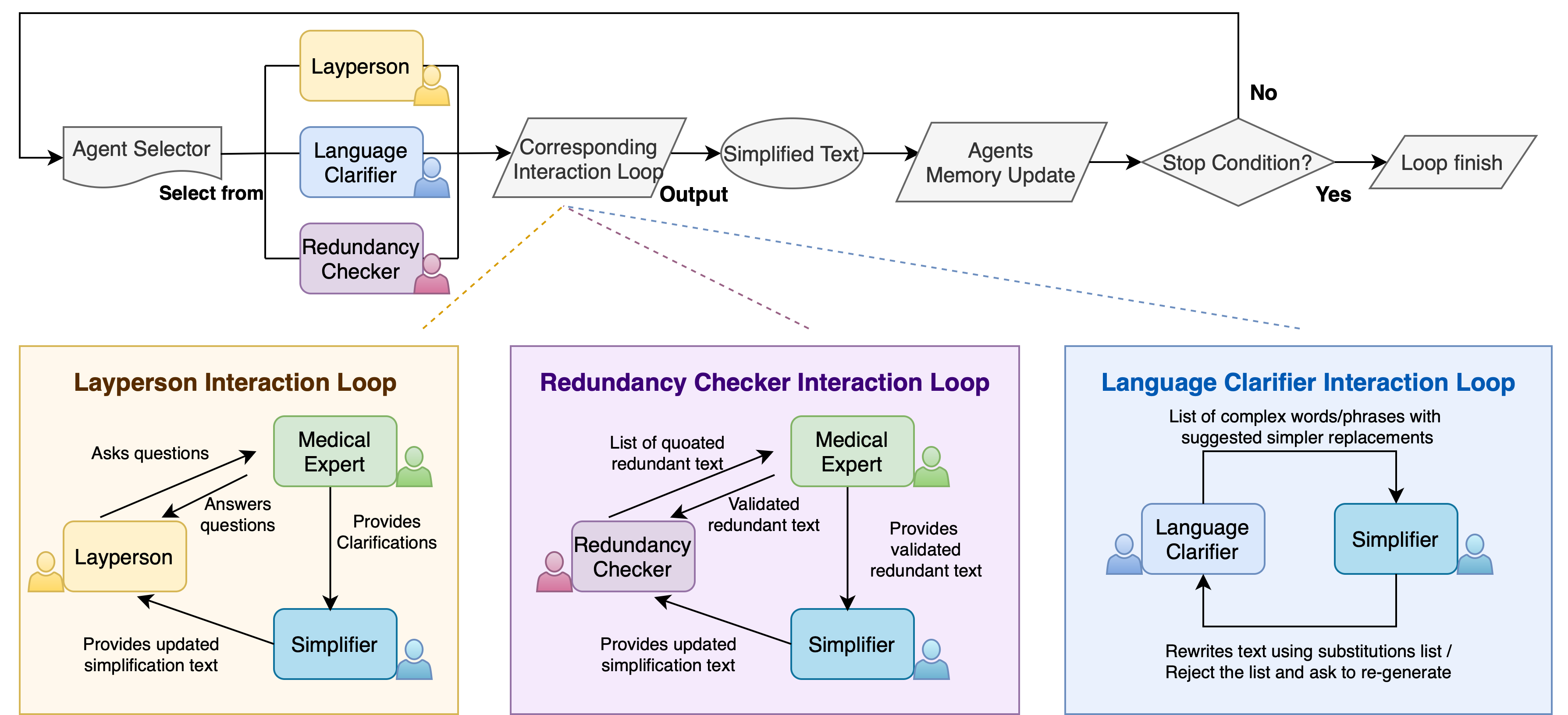}  
\caption{Society of Medical Simplifiers framework. Details of the interaction loops are presented at the bottom.}
\label{fig:framework}
\end{figure*}

Inspired by Marvin Minsky's "society of mind" (SOM) philosophy \cite{minsky1988society}, where intelligence emerges through the interaction of specialized modules, we see LLMs as ideal models for an agent-based system to tackle the complex task of medical text simplification.
SOM encourages the cooperative interaction of specialized agents in tasks requiring debate and collaboration. Prior work has shown that hierarchical agent frameworks and iterative, multi-agent discussions \cite{zhu-etal-2021-topic, wu2023autogen, chen2023reconcile, du2023improving} significantly improve the performance of natural language tasks. However, applying this multi-agent LLM-based SOM approach to medical text simplification remains largely unexplored.

To address this research gap, we introduce a novel framework for medical text simplification, grounded in SOM principles: the \textit{\textit{Society of Medical Simplifiers}}. To the best of our knowledge, this is the first multi-agent LLM system designed for medical text simplification, as well as the first to introduce the concept of interaction loops within the LLM-based SOM framework. Our five-agent framework decomposes the task into five specialized roles, i.e., \textit{Layperson, Simplifier, Medical Expert, Language Clarifier}, and \textit{Redundancy Checker}, and through iterative interaction loops, the system incrementally simplifies complex medical texts. These interaction loops enable the agents to collaborate dynamically, progressively improving the text over multiple iterations, while maintaining the integrity and accuracy of the original content. Experiments on the Cochrane dataset show that our framework outperforms current state-of-the-art methods at readability, even with a fixed number of iterations, demonstrating the potential of multi-agent LLM-based systems in simplifying complex medical texts.

The contribution of this paper can be summarized as follows:

\begin{itemize} 
    \item We introduce the first LLM-based five-agent framework for medical text simplification, decomposing the task into five specialized agent roles: Layperson, Simplifier, Medical Expert, Language Clarifier, and Redundancy Checker. 
    \item We propose the novel concept of organizing LLM agents into three  interaction loops, enabling progressive simplification while maintaining content accuracy and integrity. 
    \item Experimental assessments on the Cochrane dataset demonstrate that our framework, even with a fixed number of iterations, surpasses state-of-the-art methods in readability. \end{itemize}

\begin{figure*}[ht]
    \centering
    \small
    \subfigure[Readability metrics for Layperson and Redundancy Checker]{\includegraphics[width=0.32\textwidth]{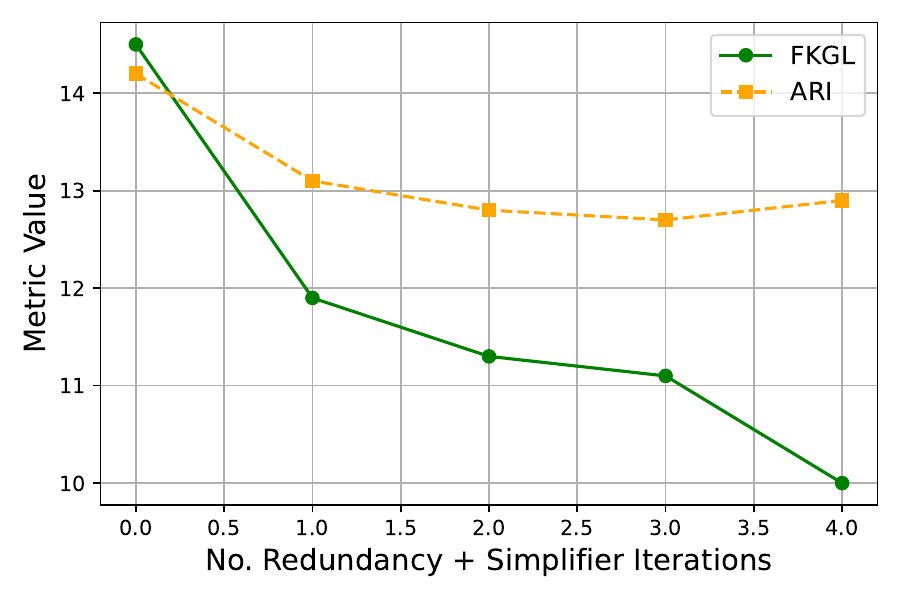}} 
     \hfill
    \subfigure[SARI DELET Metric for the Layperson and Redundancy Checker]{\includegraphics[width=0.32\textwidth]{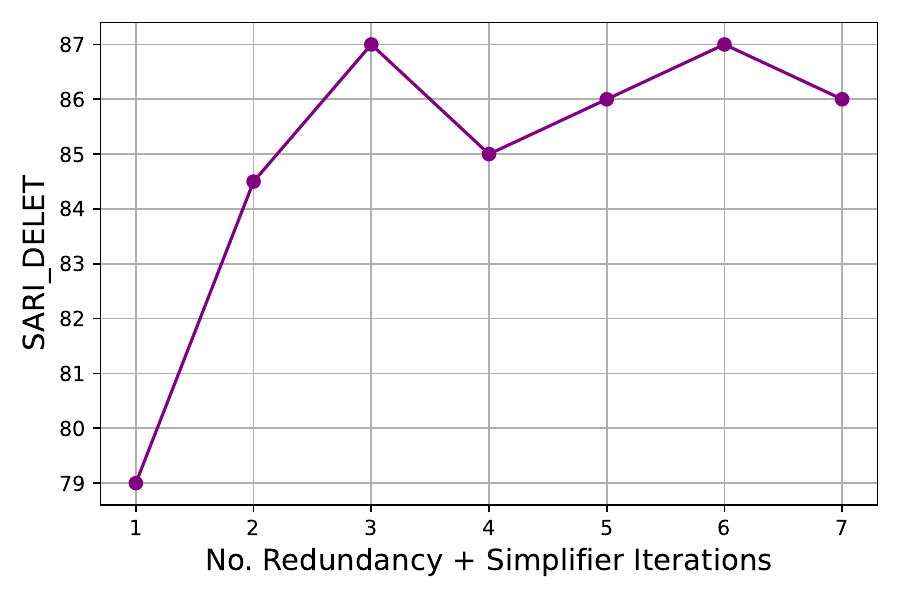}}
     \hfill
    \subfigure[Readability metrics for the Layperson and Language Clarifier]{\includegraphics[width=0.32\textwidth]{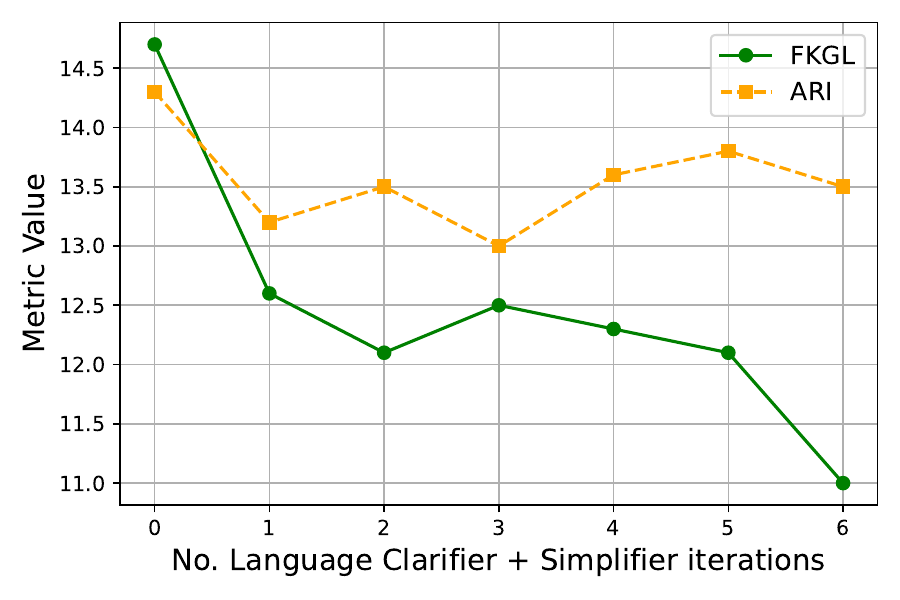}}
    \caption{Relationship between performance on metrics and the number of evaluation iterations.}
    \label{fig:scatter}
    \vspace{-10pt}
\end{figure*}

\section{Methodology}

\subsection{Agent Roles}
Figure \ref{fig:framework} provides an overview of the \textit{Society of Medical Simplifiers} framework. Five agents are assigned distinct roles that together complement the medical text simplification process. The agent roles are defined as follows:

\begin{itemize}
    \item \textbf{Layperson}: Acts as a non-expert reader, identifying complex medical jargon and posing questions to highlight areas that need simplification. This agent focuses on domain-specific content, prompting other agents to rewrite difficult medical text.
    \item \textbf{Medical Expert}: Provides detailed answers to the Layperson Agent's questions, offering clarifications that help maintain the original text's core ideas while making it more comprehensible.
    \item \textbf{Simplifier}: Uses feedback from the Medical Expert and other agents to edit and simplify the text, ensuring it remains clear and aligned to the original meaning.
    \item \textbf{Language Clarifier}: Focuses on reducing lexical complexity by identifying and suggesting simpler alternatives for non-medical terms. The suggestions are provided as a list passed to the Simplifier Agent to react on.
    \item \textbf{Redundancy Checker}: This agent identifies and recommends the removal of non-essential content. It is prompted to produce a list of redundant phrases or sentences with justifications for their exclusion.
\end{itemize}

We name Layperson, Redundancy Checker and Language Clarifier as \textbf{\textit{lead agents}}, since they are the agents driving the simplification processes. On the other hand, we refer to Medical Expert and Simplifier as \textbf{\textit{function agents}} as they act passively on leader agents' actions and perform the basic functions in this framework. Prompts used to define the agent roles are listed in \ref{sec:appendix}.

\subsection{Interaction Loops}
As illustrated at the bottom part in Figure \ref{fig:framework}, we define three distinct combinations of agents, referred to as \textit{interaction loops}. Each interaction loop consists of one lead agent and one or more function agents. These interaction groups form the core structure of the simplification process. When a function agent is selected, it enters the corresponding loop, engaging in iterative conversations with other agents until a condition is met and a simplified text is outputted. The loop is then completed.

\noindent \textbf{Layperson loop:} In the Layperson interaction loop, the Layperson first identifies difficult medical content by generating questions, to which the Medical Expert provides clarifying responses. The Simplifier agent then processes these clarifications by executing a Chain-of-Thought (CoT) prompt \cite{wei2022chain} to modify and incorporate the clarifications into the simplified text. This involves reasoning about how each clarification can be coherently integrated and rewriting it into a simpler form. The loop then ends, yielding the updated text with the clarifications. This loop performs similar function to the ATS task Complex Word Identification and Substitution as defined in previous works \cite{finnimore-etal-2019-strong, saggion-etal-2022-findings}.

\noindent \textbf{Language Clarifier Loop:} When the Language Clarifier acts as the lead agent, it starts with generating a list of complex words/phrases and their simpler replacements. The Simplifier then reviews these substitutions, deciding whether to accept or reject them. If accepted, the text is updated accordingly. If not, the Language Clarifier will revise the substitutions until they are accepted and incorporated into the text. This loop performs a function similar to the ATS task of Complex Word Identification and Substitution \cite{nisioi-etal-2017-exploring}, as outlined in prior studies. 

\noindent \textbf{Redundancy Loop:} Similarly, in its loop, the Redundancy Checker generates a list of redundant text by quoting sections of the simplified text. The Medical Expert reviews each entry to ensure no essential medical information is removed, justifying whether the text is truly redundant. Once validated, the Simplifier removes the redundant text from the document, ensuring key medical details remain intact. This loop functions similarly to a Sentence Compression module, as described in previous studies \cite{boudin2013keyphrase, shang-etal-2018-unsupervised}, where non-essential content is removed to reduce the length of the text while maintaining key facts and grammatical correctness.

\subsection{Pipeline of the Framework}
To maintain state preservation across simplification iterations, agent memories are stored as natural language. Since the Layperson's role is limited to identifying and querying complex domain-specific language, retaining previous states is unnecessary for this agent. Therefore, only the remaining four agents are equipped with memory.

At the beginning of the framework pipeline, a planning component that we refer to as the Agent Selector determines the most appropriate lead agent to act next based on the entire conversation history. The LLM-powered Agent Selector is prompted with the predefined roles of each agent and past conversations. Once a lead agent is selected, it engages with the function agents in the corresponding interaction loop until all agents agree to conclude the conversation. Each interaction loop outputs a new version of the simplified text, which is passed to the lead agents as a memory update. The \textit{logger} then updates the conversation history, and the Agent Selector chooses the next lead agent for the subsequent loop. This process continues until the predefined stop condition is reached, discussed in detail in the following.

\begin{table*}[ht]
    \centering
    \setlength{\tabcolsep}{5pt}
    \resizebox{\textwidth}{!}{%
\begin{tabular}{lcccccccc}
\toprule
\textbf{Method}                               & \textbf{SARI ↑} & \textbf{FKGL ↓} & \textbf{ARI ↓} & \textbf{BLEU ↑} & \textbf{ROUGE1 ↑} & \textbf{ROUGE2 ↑}  \\ \Xhline{1pt}
BART-UL \cite{devaraj2022evaluating}                & 40.00 & 11.97 & 13.73 & 7.90 & 38.00 & 14.00        \\ \hline
TESLEA \cite{phatak2022medical} & 40.00              & 11.84           & 13.82          & -               & 39.00             & 11.00         \\ \hline
NapSS \cite{lu2023napss}                       & \textbf{40.37}            & \textbf{10.97}  & 14.27         & \textbf{12.30 }       & \textbf{48.05}             & \textbf{19.94 }         \\ \hline
Society of Medical Simplifiers (Our)           & 40.04                  & 11.40           & \textbf{12.81}          & 8.40            & 28.03             & 9.61           \\ \bottomrule

\end{tabular}%
}
\caption{Performance metrics across various text simplification methods}
\vspace{-10pt}
\label{fig:Evaluation}
\end{table*}

\section{Experiment}
We deployed multiple instances of GPT-3.5-Turbo-1106\footnote{\url{https://platform.openai.com/docs/models/gpt-3-5-turbo}} to serve as the agents in our framework. Evaluations were conducted using the Cochrane Medical Text Simplification Dataset \cite{cochrane_dataset} sourced from the Cochrane library, which is a benchmark for medical text simplification tasks providing human-generated pairs of biomedical abstracts and their simplified version. 

It is worth noting that the focus of this experimental assessment is not on using the latest LLM, but rather on exploring and validating the effectiveness of the proposed multi-agent framework. As such, our experiments are designed to test the system's ability to improve medical text simplification through interaction loops between specialized agents, rather than to benchmark specific model performances.

\subsection{Preliminary Analyses on the Fixed Number of Iterations}

To optimize the performance of our framework while managing computational costs, we set a fixed number of iterations as the stop condition for the entire pipeline. This means the framework halts once a predefined iteration count is reached.

To determine the optimal setting for the iteration number, we initially experimented using only the basic interaction loop led by the Layperson agent, without involving the other loops. This approach allowed us to control the hyperparameter more effectively to later observe the performance of the Redundancy Checker and Language Clarifier in isolation. We tested iteration counts between 1 and 3 for the Layperson-led loop, as shown in Table \ref{fig:iter}. The experiment with 2 iterations yielded the best overall results, leading on three out of six metrics. While 3 iterations achieved the highest readability scores, it performed significantly worse on SARI scores. Therefore, we fixed the iteration count to two for the Layperson-led loop.

\begin{table}[h]
\centering
\resizebox{\columnwidth}{!}{%
\begin{tabular}{ccccccc}
\toprule
\textbf{Iteration} & \textbf{SARI ↑} & \textbf{KEEP ↑} & \textbf{DEL ↑} & \textbf{ADD ↑} & \textbf{FKGL ↓} & \textbf{ARI ↓} \\ \hline
1      & 40.00           & \textbf{28.12}  & 86.92          & 4.97          & 12.14           & 13.87          \\ 
2     & \textbf{40.32}  & 27.64           & \textbf{87.73} & \textbf{5.59}  & 12.08           & 13.22          \\ 
3      & 38.28           & 24.13           & 86.44          & 4.26         & \textbf{11.93}  & \textbf{12.85} \\ 
\bottomrule
\end{tabular}%
}
\caption{Performance metrics across different numbers of iterations for Layperson Loop}
\label{fig:iter}
\end{table}

For the Redundancy Checker and Language Clarifier loops, our focus shifts to readability metrics only because these agents are primarily responsible for improving text readability. We ran additional experiments with varying iteration counts for combinations of Layperson and Redundancy Checker, and Layperson and Language Clarifier. We additionally recorded the SARI DELETE component of the SARI metric for the Redundancy Checker to evaluate the text removal accuracy for the redundant parts. Results shown in Figure \ref{fig:scatter} indicate a negative correlation between iteration count and readability for the Redundancy Checker, with its highest SARI DELETE score occurring at three iterations. However, for both agents, the most significant gains were observed by the second iteration, after which improvements level out. Thus, we also selected two iterations as the fixed setting for the Redundancy Checker and Language Clarifier.

\subsection{Result and Discussion}
To evaluate the effectiveness of our proposed framework, we ran the experiments on the Cochrane simplification dataset and presented the results along with the recent literature in Table \ref{fig:Evaluation}. Following the conclusion of the previous optimal iteration number investigation, we adopted 2 as the fixed number of loops. This means that each interaction loop will be entered twice by the corresponding lead agent. The whole framework will stop running once all interaction loops have been selected two times and the Agent Selector is out of options.

As shown in Table \ref{fig:Evaluation}, the designed framework outperforms state-of-the-art methods on the ARI readability metric and also demonstrates superior performance on SARI and FKGL compared to most existing approaches. However, the framework shows a noticeable decrease in ROUGE scores. This issue could stem from excessive content added by the Medical Expert or the removal of relevant information by the Redundancy Checker. Overall, while further improvements are necessary to balance content preservation and simplification, the experiments demonstrate the current framework as a competitive and state-of-the-art approach.

\section{Conclusion}
In this paper, we introduced the Society of Medical Simplifiers, a novel multi-agent LLM-based framework for medical text simplification. Inspired by the SOM philosophy, our framework designs and organizes five specialized agents into iterative interaction loops, enabling collaborative simplification of complex medical texts. Our experiments on the Cochrane dataset show that the Society of Medical Simplifiers outperforms existing methods in terms of readability and simplification.

\section{Limitations} 
Future improvements to our framework include evaluating its performance on a wider family of LLMs, such as the Llama 3 models \cite{llama3}, the GPT-4 models \cite{gpt4}, and Mi(x)tral \cite{mixtral}. Additionally, experimenting with a larger number of iterations, and potentially automating the selection of optimal iteration counts through LLM inference, could further boost performance. We also aim to explore more complex interactions between agents, introducing new roles that emphasize preserving the original context while simplifying the text. 

\section*{Acknowledgements}

This work was supported by Sulis, the Tier-2 high-performance computing facility at the University of Warwick, funded by the EPSRC through the HPC Midlands+ Consortium (EPSRC grant no. EP/T022108/1). 


\bibliography{custom}

\newpage

\appendix

\setcounter{table}{0}
\renewcommand{\thetable}{A\arabic{table}}
\setcounter{figure}{0}
\renewcommand{\thefigure}{A\arabic{figure}}

\section{Agent Role Prompts}
\label{sec:appendix}
We present the prompts we used to define all five agents roles below.

\subsection{Layperson}
\begin{lstlisting}
You are a casual person who is reading a complicated medical text.
You are confused by the medical jargon, unfamiliar terms and numerical information, making it difficult to understand the key takeaways.
You are in a room with a medical expert and simplifier agent. 
                
You must ask at least 4 questions about the text, to the medical expert, who will clarify terms, conclusions, concepts or sections which you don't understand.
                
Possible questions:
    - Can you explain X?
    - I don't understand X.
    - What are the main takeaways or key points?
    - How does X work, and what are its implications?
    - What are the potential risks or side effects associated with X?
                    
You must output a numbered list of questions.
The simplifier agent will produce an updated version of the text to meet your needs.
\end{lstlisting}

\subsection{Medical Expert}
\begin{lstlisting}
You are a medical expert.
You are in a room with a casual person and a simplifier agent.
                
You will help a casual person understand a complicated medical abstract by answering their questions and providing clarifications in a simplified form.
Your advice will help the simplifier edit the text to satisfy the casual person.
Ensure your answers restate the context of the question.
Your answers must be brief, using as little words as possible.
After the text has been rewritten, review it to check if it is medically accurate, potentially outputting a list of comments. If it is accurate, state so.
\end{lstlisting}

\subsection{Simplifier}
\begin{lstlisting}
You are a simplifier who is in a conversation with a casual person who does not understand a complex medical text and will ask questions about the text.

A medical expert will answer their questions. You must rewrite the original medical text in a simplified form. Your messages in the conversation must be your latest simplified version of the entire medical text from your memory.
You should title this "Latest Simplification" and ask the casual person for further questions. 
You can simplify clarifications from the medical expert.
You must not answer questions from the casual person, or answer any medical questions.
\end{lstlisting}

\subsection{Language Clarifier}
\begin{lstlisting}
You are given a medical text which needs simplifying.
Identify complex words, phrases (non-medical) and sentence structures.

Suggest replacements: simpler equivalent paraphrases, using common vocabulary and minimising technical jargon.
Suggest to join, split or rearrange sentences which are too long or segmented.
Output a list of suggestions. Do not rewrite the entire text."""),
           
\end{lstlisting}

\subsection{Redundancy Checker}
\begin{lstlisting}
You are given a simplified version of a medical text.
Identify redundant phrases or terms which hurt clarity and do not add to the key information of the text, and should be removed.

Parts of the text must each be very short - 5 words maximum - to remove.    
Medical related information must not be removed as this is essential. Only remove text which is completely unrelated to medicine.
Remove very little information from the text.
Output a list of comments quoting short redundant parts, with a very brief description. If there is no text to remove, state so. 
\end{lstlisting}

\end{document}